\documentclass[a4paper]{article}

\usepackage{times}
\usepackage{epsfig}
\usepackage{graphicx}
\usepackage{amsmath}
\usepackage{amssymb}

\usepackage{multirow}
\usepackage{bigstrut}
\usepackage{comment}
\usepackage{bm}
\usepackage[ruled,linesnumbered]{algorithm2e}
\usepackage{color}
\usepackage{url}
\usepackage[pagebackref,breaklinks,colorlinks]{hyperref}

\usepackage{ISCSLP2022}

\title{The Conversational Short-phrase Speaker Diarization (CSSD) Task: Dataset, Evaluation Metric and Baselines}

\name{Gaofeng Cheng$^{1,\dagger,\ddagger}$\thanks{$^\dagger$ Equal contribution.}\thanks{$^\ddagger$ Corresponding author.}, Yifan Chen$^{1,2,\dagger}$, Runyan Yang$^{1,2}$,Qingxuan Li$^{3}$, Zehui Yang$^{1,2}$, Lingxuan Ye$^{1,2}$, Pengyuan Zhang$^{1,2}$,Qingqing Zhang$^{4}$, Lei Xie$^{5}$, Yanmin Qian$^{6}$,Kong Aik Lee$^{7}$, Yonghong Yan$^{1,2}$\thanks{This work was partially supported by the Youth Innovation Promotion Association, Chinese Academy of Sciences and the Frontier Exploration Project Independently Deployed by Institute of Acoustics, Chinese Academy of Sciences under Grant QYTS202011.}}

\address{
  $^1$Key Laboratory of Speech Acoustics and Content Understanding, Institute of Acoustics, \\ Chinese Academy of Sciences, China,
  $^2$University of Chinese Academy of Sciences, China\\
  $^3$Tsinghua University, Beijing, China,
  $^4$Magic Data Technology Co., Ltd., China\\
  $^5$Northwestern Polytechnical University, China,
  $^6$Shanghai Jiao Tong University, Shanghai, China\\
  $^7$Institute for Infocomm Research, A*Star, Singapore}
  
\email{\{chenggaofeng,chenyifan\}@hccl.ioa.ac.cn}

\begin{document}

\maketitle
\begin{abstract}

The conversation scenario is one of the most important and most challenging scenarios for speech processing technologies because people in conversation respond to each other in a casual style. Detecting the speech activities of each person in a conversation is vital to downstream tasks, like natural language processing, machine translation, etc. People refer to the detection technology of ``who speak when'' as speaker diarization (SD). Traditionally, diarization error rate (DER) has been used as the standard evaluation metric of SD systems for a long time. However, DER fails to give enough importance to short conversational phrases, which are short but important on the semantic level. Also, a carefully and accurately manually-annotated testing dataset suitable for evaluating the conversational SD technologies is still unavailable in the speech community. 
In this paper, we design and describe the Conversational Short-phrases Speaker Diarization (CSSD) task, which consists of training and testing datasets, evaluation metric and baselines. In the dataset aspect, despite the previously open-sourced 180-hour conversational MagicData-RAMC dataset, we prepare an individual 20-hour conversational speech test dataset with carefully and artificially verified speakers timestamps annotations for the CSSD task. In the metric aspect, we design the new conversational DER (CDER) evaluation metric, which calculates the SD accuracy at the utterance level. In the baseline aspect, we adopt a commonly used method: Variational Bayes HMM x-vector system, as the baseline of the CSSD task. 
Our evaluation metric is publicly available at  \url{https://github.com/SpeechClub/CDER_Metric}.


\end{abstract}

\section{Introduction}

The conversation scenario is one of the most important and, at the same time, most challenging scenarios for speech processing technologies because people in conversation respond to each other in a casual style and continue the dialog with coherent questions and opinions instead of stiffy answering each other’s questions. Detecting the speech activities of each person in a conversation is vital to downstream tasks, such as speech recognition, natural language processing, machine translation, etc~\cite{park2022review,cheng2022eteh}. People refer to the detection technology of ``who speaks when'' as speaker diarization. Speaker diarization aims at detecting the speech activities of each person in a conversation, and it has been extensively studied in rencent years~\cite{anguera2012speaker}. 

Traditionally, the evaluation metric of speaker diarization systems is diarization error rate (DER), which is calculated as the summed time of three different errors of speaker confusion (ERR), false alarm (FA), and missed detection (MISS) divided by the total time duration. Although DER has been used as the standard evaluation metric for speaker diarization for a long time, it fails to give enough importance to the short conversational phrases, which last for a short time but play an important role on the semantic level. Moreover, the speech community lacks an evaluation metric that gives enough emphasis on the diarization accuracy of short phrases in conversation.

Therefore, we have released the open-source MagicData-RAMC~\cite{yang2022open}, consisting of 180 hours of conversational speech data recorded from native speakers of Mandarin Chinese over mobile phones with a sampling rate of 16 kHz. Apart from the already published MagicData-RAMC corpus, we prepare an individual 20 hours conversational speech with artificially verified annotations. Moreover, we propose the Conversational Short-phrase Speaker Diarization Challenge (CSSD) as an ISCSLP 2022 challenge. For the challenge, we design a new accuracy evaluation metric, which calculates the speaker diarization accuracy at the utterance level. We provide a detailed introduction of the dataset, rules, evaluation methods, and baseline systems, aiming to promote reproducible research in this field further.

\section{Related Work}

\subsection{Speaker Diarization Dataset}

Speech data is essential for data-driven spoken language processing methods~\cite{graves2014towards, yang2021keyword, guo2021far, Dong2018SpeechTransformerAN}. For speaker diarization task, there are some multi-speaker datasets collected in diverse scenarios, e.g. meeting scenario and dialog scenario. For meeting scenario, CHIL \cite{mostefa2007chil} and AMI \cite{ami} corpora are composed of audio recorded during the conferences held in academic laboratories. AISHELL-4 \cite{fu2021aishell} is a 120-hour Mandarin speech dataset collected in conference scenarios. The number of attendees in each session is between four and eight. The indoor conversation corpus CHiME-6 \cite{watanabe2020chime} consists of unsegmented recordings of twenty separate dinner-party conversations among four friends captured by Kinect devices. It has a more casual and natural speaking style yet relatively low recording quality. For dialog scenario, Switchboard \cite{godfrey1992switchboard} is a classic dataset of English telephony conversations with similar settings and different scales. HKUST is a Mandarin conversational corpus \cite{liu2006hkust} made up of spontaneous telephone conversations. Audios in these three telephony conversational datasets are recorded with a sampling rate of 8 kHz, which is incompatible with the demand of some speech processing systems nowadays. 


\subsection{Evaluation Metric for Speaker Diarization}
\label{sec::relate_metric}
The most commonly used metric for speaker diarization is DER~\cite{park2022review}, which is introduced for the NIST Rich Transcription Spring 2003 Evaluation (RT-03S). Specifically, DER is defined as:
\begin{equation}
    DER=\frac{FA+MISS+ERROR}{TOTAL}
\end{equation}
where $FA$ is the total hypothesis speaker time not attributed to a reference speaker, $MISS$ is the total reference speaker time not attributed to a hypothesis speaker, $ERROR$ is the total reference speaker time attributed to the wrong speaker, and $TOTAL$ is the total reference speaker time, i.e. the sum of the duration of all reference speaker segments~\cite{ryant2018first}. In most challenges and studying, a 0.25 second of collar is set around each segmentation boundaries, which aims at mitigating  annotation error of reference transcript~\cite{park2022review}. The calculation of DER is following.

\begin{algorithm}[htbp]
	\footnotesize
    \caption{The Calculation of DER.}
    \label{algo:DER}
    \KwIn{reference speaker timestamps $\{S_i,U_i\}_{i=1}^L$ , $S_i$ represents speaker label, $U_i=\{[T_{Start,1},T_{End,1} ],\dots,[T_{Start,j},T_{End,j} ],\dots\}$ represent timestamp of start and end of each utterance, L is the number of speakers), hypothesis speaker timestamp $\{S_i',U_i'\}_{i=1}^M$}
    Initialize: $T_{ERR}=0,T_{FA}=0,T_{MISS}=0,T_{TOTAL}=0$ \\
    Find matching between reference speakers $\{S_i\}_{i=1}^L$ and hypothesis speakers $\{S_i^{'}\}_{i=1}^M$
    
    \For{$i=1,2,\dots,L$}
    {
        Calculate $t_{ERR},t_{FA},t_{MISS},t_{TOTAL}$ of speaker $S_i$. \\
        $T_{ERR}=T_{ERR}+t_{ERR},T_{FA}=T_{FA}+t_{FA},$ \\
        $T_{MISS}=T_{MISS}+t_{MISS},$\\ 
        $T_{TOTAL}=T_{TOTAL}+t_{TOTAL}$ \\
    }
    $DER=(T_{ERR}+T_{FA}+T_{MISS})/T_{TOTAL}$
\end{algorithm}

Besides, Jaccard error rate (JER) is introduced by~\cite{ryant2019second}, which is developed as a metric for DIHARD II. Inspired by the evaluating method in image segmentation area, JER uses Jaccard similarity index that is the ratio between the sizes of the intersections and unions of two sets of segments. 



However, all these evaluation metrics fail to give enough importance to the short conversational phrases. For comparison, in our proposed evaluation metric, Conversational Diarization Error Rate (CDER), all type of mistakes are equally reflected in the final evaluation metric, regardless of the length of the spoken sentence.

\subsection{Speaker Diarization under Conversation Scenario}

Recent speaker diarization systems could be devided into two types according to whether to use accurate timestamp annotations, i.e. clustering based approaches and fully supervised approaches~\cite{anguera2012speaker}.  Clustering-based methods generate diarization results in an unsupervised way. A typical procedure is firstly conducting speech activity detection. Then embeddings such as x-vectors are extracting and finally clustering~\cite{snyder2018x}. Based on this, several sophisticated clustering models are designed. Fox et al.~\cite{fox2007sticky} introduce the Dirichlet processes mixture model into the diarization task. Diez et al.~\cite{diez2019analysis} designed a Bayesian model, i.e. VB-HMM, to simulate conversational state. And graph neural networks are used for clustering by Wang et al.~\cite{wang2020speaker} to model the conversation of multi-speakers. Besides, semantic information are introduced to assist speaker embeddings clustering by using word boundary information~\cite{park2020speaker} and utilizing endpoints of sentences~\cite{shafey2019joint}. However, clustering-based method does not optimized diarization object directly~\cite{chen2022interrelate} and the basic assumption of clustering-based method is that a chunk belongs only to one person, which restricts the capacity of handling overlapped conversation. 

Fully supervised approaches need data with accurate timestamp annotation, thus optimizing diarization objects in a fully supervised way. And it could achieve promising performance, especially when speech from different speakers is highly overlapped~\cite{fujita2019end,medennikov2020target}. UIS-RNN~\cite{zhang2019fully} uses the recurrent neural networks (RNN) to model the transition probabilities among different speakers in conversation, which could output diarization results in an online fashion. EEND~\cite{fujita2019end,maiti2021end} discards the use of the speakers' embeddings and directly optimizes the diarization task in an end-to-end manner. Based on EEND, EDA-EEND~\cite{horiguchi2020end} decodes the diarization results of different speakers recursively, in order to tackle the challenge of an uncertain number of speakers. Maiti et al.~\cite{maiti2021end} handle this challenge by using the variable-number permutation-invariant training. Xue et al.~\cite{xue2021online} introduce a speaker-tracing buffer in order to solve the across-chunk permutation issue when extending the EEND to online conversation scenario. All these methods treat speaker diarization as a sequence-to-sequence problem, aiming to decode a sequence of speaker labels from an initial sequence of features.  Another perspective is to regard speaker diarization as a detection task, where the goal is to predict the position of each person in a conversation. RPNSD~\cite{huang2020speaker} utilizes the Faster R-CNN to detect activities of each person in conversation. And TS-VAD~\cite{medennikov2020target} compares the embedding of a chunk of audio with the embedding of a target speaker to detect whether the corresponding person speaks or not.

\section{Datasets}

\subsection{MagicData-RAMC}

MagicData-RAMC~\cite{yang2022open} contains 180 hours of dialog speech in total. The dataset is divided into 149.65 hours training set, 9.89 hours development set, and 20.64 hours test set (not the final evaluation set of the CSSD challenge), consisting of 289, 19, and 43 conversations, respectively. The original partition of the speech data is provided in TSV format and the 180-hour data will be released to the participants at the beginning of the challenge. Each conversation is of 30.80 minutes duration on average. The numbers of participants involved in three subsets are 556, 38, and 86, respectively. The gender and region distribution is roughly proportional to the entire dataset. Table~\ref{tab:partition} provides a summary of the partition of the corpus. It is recommended to provide a result based on MagicData-RAMC’s test set which helps to promote the reproducibility of the research.
The dataset is collected indoors. The domestic environments are small rooms under 20 square meters in area, and the reverberation time (RT60) is less than 0.4 seconds. The environments are relatively quiet during recording, with ambient noise level lower than 40 dB. The audios are recorded over mainstream smartphones, including Android phones and iPhones. The ratio of Android phones to iPhones is around 1:1. All recording devices work at 16 kHz, 16-bit to guarantee high recording quality. 

\begin{table}[htbp]
  \caption{Corpus partition}
  \label{tab:partition}
  \centering
  \begin{tabular}{l c c c}
    \toprule
     & \textbf{Training} & \textbf{Development} &\textbf{Test}\\
    \midrule
    Duration (h) & $149.65$ & $9.89$    & $20.64$         \\
    \#Sample & $289$  & $19$  & $43$             \\
    \#Speaker  & $556$  & $38$  &  $86$   \\
    \#Male  & $307$  & $23$    &  $49$ \\
    \#Female & $249$  & $15$    &   $37$   \\
    \#Northern & $271$ & $20$ & $52$\\
    \#Southern & $285$ & $18$ & $34$\\
    \bottomrule
  \end{tabular}
  \vspace{-0.1in}
\end{table}

There are a total of 663 speakers involved in the recording, of which 295 are female and 368 are male. Each segment is labeled with the corresponding speaker-id. All participants are native and fluent Mandarin Chinese speakers with slight variations of accent and participants in each group are acquaintances. The accent region is roughly balanced, with 334 Northern Chinese speakers and 329 Southern Chinese speakers. Besides, each speaker participates in up to three conversations. 
All speech data are manually labeled. Sound segments without semantic information during the conversations, including laughter, music, and other noise events, are annotated with specific symbols. Phenomena common in spontaneous communications, such as colloquial expressions, partial words, repetitions, and other speech disfluencies, are recorded and fully transcribed. The precise voice activity timestamps of each speaker are provided. All transcriptions of the speech data are prepared in TXT format for each dialog. The start and end times of all segments are specified to within a few milliseconds. The statistics are presented in Table~\ref{tab:speech}. Topics are freely chosen by the participants and the conversations in MagicData-RAMC are classified into 15 diversified domains, ranging from science and technology to ordinary life.

\begin{table}[htbp]
  \caption{Statistics of speech}
  \label{tab:speech}
  \centering
  \begin{tabular}{l c c c}
    \toprule
    \textbf{Statistical Criterion} & {\textbf{Max}} & {\textbf{Min}} & {\textbf{Average}} \\
    \midrule
    Sample Duration (min)   & $33.02$   & $14.06$   & $30.80$   \\
    \#Segments Per Sample   & $1215$    & $231$     & $624.86$  \\
    Segment Duration (s)    & $14.91$   & $0.09$    & $2.54$    \\
    \#Tokens Per Segment    & $89$      & $1$       & $13.58$   \\
    \#Segments Per Speaker  & $1155$    & $46$      & $304.55$  \\
    \bottomrule
  \end{tabular}
  \vspace{-0.1in}
\end{table}

The test set contains 41 conversations, which amount to 20 hours. The number of participants is 82 with 43 males and 39 females. There are 9 conversations between women, 11 between men and 21 between the opposite sex.

\subsection{CSSD Testset}

Despite the MagicData-RAMC dataset, we prepare an individual 20-hour conversational speech test dataset with artificially verified speakers timestamps annotations for the CSSD task. Each segment is labeled with the corresponding speaker-id. Besides, all participants are native Mandarin Chinese speakers with slight variations of accent and participants in each group are acquaintances. The accent region and gender is roughly balanced. There are a total of 82 speakers involved in the recording, of which 39 are female and 43 are male. There are two speaker in each conversation, and 41 conversations are recorded. Among the 41 conversations, the number of conversations in which speakers consists of a man and a woman is 21. And the number of conversations between women is 9, while the number of conversations between men is 11.

\subsection{Other Datasets}

Apart from speaker diarization datasets that could be used in fully supervised methods, there are several speaker recognition datasets needed in clustering-based methods, i.e. CN-Celeb~\cite{li2022cn} and VoxCeleb~\cite{chung2018voxceleb2}. CN-Celeb includes two large-scale Chinese speaker recognition corpora. CN-Celeb1 consists of 274 hours audio from 1000 speakers, while CN-Celeb2 contains 1090 hours audio from 2000 speakers. These datasets could be used to train the speaker embedding extractor.

\section{Evaluation Metric}

\begin{algorithm}[!h]
	\footnotesize
    \caption{The Calculation of CDER.}
    \label{algo:CDER}
    \KwIn{reference speaker timestamp  $\{S_i,U_i\}_{i=1}^L$ ($S_i$ represents speaker label, $U_i=\{[T_{Start,1},T_{End,1} ],\dots,[T_{Start,j},T_{End,j} ],\dots\}$ represent timestamp of start and end of each utterance, $L$ is the number of speakers); hypothesis speaker timestamp $\{S_i',U_i'\}_{i=1}^M$; matching error threshold $\eta(0<\eta<1)$.}

    Initialize: $N_{ERROR}=0,N_{TOTAL}=0$, which means the number of mistakes and the number of total utterances. $\widetilde{U}_{i}=\{\}$, $\widetilde{U}_{i'}^{\prime}=\{\}$, which stores the merged timestamps from $U_{i}$ and $U_{i'}^{\prime}$\\
    Find matching between reference speaker $\{S_i\}_{i=1}^L$ and hypothesis speaker $\{S_i'\}_{i=1}^M$ \\
    // merge the utterances from the same person \\
    \For{$i=1,2,\dots,L$}
    {
        $K = utt\ num\ of\ U_{i}$ \\
        Sort $U_{i}$ in chronological order \\
        $j = 1$\\
        \While{$j \leq K$}
        {
            $step = 1$ \\
            \While{$j + step \leq K$}
            {
                \eIf{other\ speaker\ is\ not\ active\ in $[T_{Start,j}, T_{End,j+step}]$:}
                {
                    $step=step+1$
                }
                {
                    break
                }
            }
            $\widetilde{U}_{i}$ = $\widetilde{U}_{i} \cup \{[T_{Start,j}, T_{End,j+step-1}]\}$ \\
            $j = j + step$
        }
    }
    Get $\widetilde{U}_{i'}^{\prime}$ in the same way \\
    
    \For{$i=1,2,\dots,L$}
    {
        $P = utt\ num\ of\ \widetilde{U}_{i}$ \\
        \If{$S_i$\ is\ not\ matched\ to\ a\ hypothesis\ speaker:}
        {
            $N_{ERROR}=N_{ERROR}+P$\\
            continue \\
        }
        Assume $S_i$ is matched to hypothesis speaker $S_{i'}'$ \\

        \For{$j=1,2,\dots,P$}
        {
            Match the $j$-th reference utterance of $\widetilde{U}_{i}$ : $R=[T_{Start,j},T_{End,j}]$, to a hypothesis utterance of $\widetilde{U}_{i'}^{\prime}$ : $H=[T_{Start,j'}',T_{End,j'}']$.
            $N_{TOTAL}=N_{TOTAL}+1$ \\
            \If{$\frac{|R\cap H|}{|R\cup H|} < \eta$:}
            {
                $N_{ERROR}=N_{ERROR}+1$
            }
        }

    Assume $E$ is the number of hypothesis utterances of $\widetilde{U}_{i'}^{\prime}$ not matched to a reference utterance. \\
    $N_{ERROR}=N_{ERROR}+E$
    }
    $CDER=N_{ERROR}/N_{TOTAL}$
\end{algorithm}

As mentioned in Section~\ref{sec::relate_metric}, the performance of speaker diarization systems is generally measured by DER and so on. These evaluation metrics calculate the percentage of reference speaker time that is not correctly attributed to a speaker~\cite{ryant2018first}, and could reasonably evaluate the overall performance of the speaker diarization system on the time duration level. However, in real conversations, there are cases that a shorter duration contains vital information. And the current speaker diarization system is not robust enough for short-term speech fragments. The evaluation of the speaker diarization system based on the time duration is difficult to reflect the recognition performance of short-term segments. Our basic idea is that for each speaker, regardless of the length of the spoken sentence, all type of mistakes should be equally reflected in the final evaluation metric. Based on this, we intend to evaluate the performance of the speaker diarization system on the sentence level under conversational scenario (utterance level). We adopt Conversational-DER (CDER) to evaluate the speaker diarization system. The CDER is defined as follows.
\begin{equation}\label{equ::cder}
    CDER=\frac{The\ number\ of\ mistakes}{The\ number\ of\ total\ utterances}
\end{equation}
A more specific algorithm that calculates CDER is shown in Algorithm~\ref{algo:CDER}.

For CDER calculation, we will firstly merge the utterances from the same person. For example, assuming $A_i$ represents the i-th utterance from speaker $A$ and $NS$ represents non-speech segments, $A_1,NS,A_2,NS,B_1,A_3,A_4,B_2,A_5,C_1$ will be merge to $A_1',B_1',A_3',B_2',A_5',C_1'$. A merged utterance ($A_1'$) would preserve the timestamps of the start time of the first utterance ($A_1$) and the end time of the last utterance ($A_2$). Then, we will match each reference utterance to a hypothesis utterance. And we will compare the reference utterance with the matched hypothesis utterance to judge whether the prediction is correct or wrong. Finally we will calculate the CDER using Equation~\ref{equ::cder}.

\section{Baseline}
Our baseline system\footnote{Our baseline system is publicly available at  \url{https://github.com/MagicHub-io/MagicData-RAMC}.} consists of three components: speaker activity detection (SAD), speaker embedding extractor, and clustering. Following the experiment setting of Variational Bayes HMM x-vectors (VBx)~\cite{landini2022bayesian}, we use a Kaldi-based SAD module for detecting speech activity. We adopt ResNet trained on VoxCeleb Dataset~\cite{chung2018voxceleb2}, CN-Celeb Corpus~\cite{li2022cn}, and the training set of MagicData-RAMC to obtain the speaker embedding extractor.

For training details, the SAD module utilizes 40-dimensional Mel frequency cepstral coefficients (MFCC) with 25 ms frame length and 10 ms stride as input features to detect the speech activity. ResNet-101 with two fully connected layers is employed to conduct speaker classification task with 64-dimensional filterbank features extracted every 10 ms with 25 ms window. The raw waveform is split every 4s (400 dimensions) to form ResNet input. We train the speaker embedding network using stochastic gradient descent (SGD) optimizer with a $0.9$ momentum factor and $0.0001$ L2 regularization factor. 

Speaker embeddings are extracted on SAD result every 240 ms, and the chunk length is set to 1.5s. Besides, probabilistic linear discriminant analysis (PLDA) is conducted to reduce the dimension of the embeddings from 256 to 128. For the clustering part, we use Variational Bayes HMM~\cite{landini2022bayesian} on this task. An agglomerative hierarchical clustering algorithm with VBx is conducted to get the clustering result. In the VBx, the acoustic scaling factor $Fa$ is set to $0.3$, and the speaker regularization coefficient is set to $17$. The probability of not switching speakers between frames is $0.99$. We present the experimental result in Table~\ref{sd_result}. 

\begin{table}[htbp]
    \caption{Speaker diarization results of VBx system}
	\label{sd_result}
	\begin{center}
		\begin{tabular}{ccccc}
			\toprule
			\multirow{2}{*}{Method} & \multirow{2}{*}{Subset} & \multicolumn{2}{c}{DER} & \multirow{2}{*}{CDER} \\ 
			\cline{3-4}
			 & & collar 0.25s & collar 0s & \bigstrut\\
			\hline
			\multirow{2}{*}{VBx} & Dev & 5.57 & 17.48 & 26.9 \bigstrut\\
			 & Test & 7.96 & 19.90 & 28.2 \\
			\bottomrule
		\end{tabular}
	\vspace{-0.1in}
	\end{center}
\end{table}

\begin{figure}[htbp]
 	\vspace{-3mm}
	\begin{center}
		\begin{minipage}[t]{0.95\linewidth}
			\includegraphics[width=1.0\linewidth]{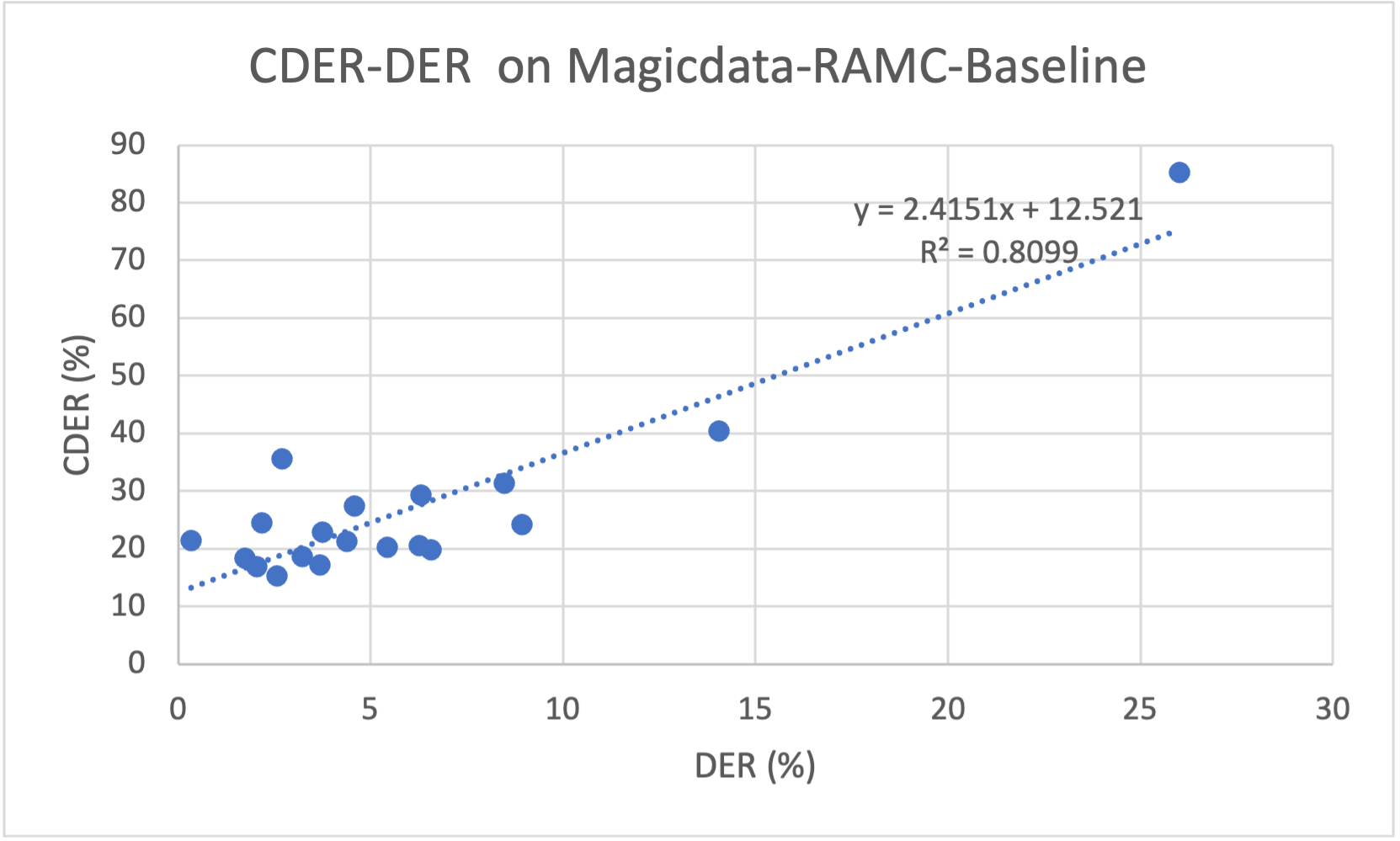}
		\end{minipage}
	\end{center}
	\vspace{-3mm}
	\caption{The correlation between CDER and DER on the speaker diarization result.}
	\label{image::corr}
\end{figure}

Besides, we calculate the correlation between CDER and DER on speaker diarization result of our baseline. As shown in Figure~\ref{image::corr}, the CDER and DER are linearly correlated.

\section{Conclusion}
In this paper, we design and describe the Conversational Short-phrases Speaker Diarization (CSSD) task, which consists of training and testing datasets, evaluation metric and baselines. In the dataset aspect, despite the previously open-sourced 180-hour conversational MagicData-RAMC dataset, we prepare an individual 20-hour conversational speech test dataset with carefully and artificially verified speakers timestamps annotations for the CSSD task. In the metric aspect, we design the new conversational DER (CDER) evaluation metric, which calculates the SD accuracy at the utterance level. In the baseline aspect, we adopt a commonly used system, i.e. Variational Bayes HMM x-vector system~\cite{landini2022bayesian}, as the baseline of the CSSD task. Besides, we evaluate our baseline's performance using the proposed CDER metric, and present the evaluation results.


\bibliographystyle{IEEEtran}

\bibliography{template}

\begin{thebibliography}{10}
\providecommand{\url}[1]{#1}
\csname url@samestyle\endcsname
\providecommand{\newblock}{\relax}
\providecommand{\bibinfo}[2]{#2}
\providecommand{\BIBentrySTDinterwordspacing}{\spaceskip=0pt\relax}
\providecommand{\BIBentryALTinterwordstretchfactor}{4}
\providecommand{\BIBentryALTinterwordspacing}{\spaceskip=\fontdimen2\font plus
\BIBentryALTinterwordstretchfactor\fontdimen3\font minus
  \fontdimen4\font\relax}
\providecommand{\BIBforeignlanguage}[2]{{%
\expandafter\ifx\csname l@#1\endcsname\relax
\typeout{** WARNING: IEEEtran.bst: No hyphenation pattern has been}%
\typeout{** loaded for the language `#1'. Using the pattern for}%
\typeout{** the default language instead.}%
\else
\language=\csname l@#1\endcsname
\fi
#2}}
\providecommand{\BIBdecl}{\relax}
\BIBdecl

\bibitem{park2022review}
T.~J. Park, N.~Kanda, D.~Dimitriadis, K.~J. Han, S.~Watanabe, and S.~Narayanan,
  ``A review of speaker diarization: Recent advances with deep learning,''
  \emph{Computer Speech \& Language}, vol.~72, p. 101317, 2022.

\bibitem{cheng2022eteh}
G.~Cheng, H.~Miao, R.~Yang, K.~Deng, and Y.~Yan, ``Eteh: Unified
  attention-based end-to-end asr and kws architecture,'' \emph{IEEE/ACM
  Transactions on Audio, Speech, and Language Processing}, vol.~30, pp.
  1360--1373, 2022.

\bibitem{anguera2012speaker}
X.~Anguera, S.~Bozonnet, N.~Evans, C.~Fredouille, G.~Friedland, and O.~Vinyals,
  ``Speaker diarization: A review of recent research,'' \emph{IEEE Transactions
  on audio, speech, and language processing}, vol.~20, no.~2, pp. 356--370,
  2012.

\bibitem{yang2022open}
Z.~Yang, Y.~Chen, L.~Luo, R.~Yang, L.~Ye, G.~Cheng, J.~Xu, Y.~Jin, Q.~Zhang,
  P.~Zhang \emph{et~al.}, ``Open source magicdata-ramc: A rich annotated
  mandarin conversational (ramc) speech dataset,'' \emph{arXiv preprint
  arXiv:2203.16844}, 2022.

\bibitem{graves2014towards}
A.~Graves and N.~Jaitly, ``Towards end-to-end speech recognition with recurrent
  neural networks,'' in \emph{International conference on machine
  learning}.\hskip 1em plus 0.5em minus 0.4em\relax PMLR, 2014, pp. 1764--1772.

\bibitem{yang2021keyword}
R.~Yang, G.~Cheng, H.~Miao, T.~Li, P.~Zhang, and Y.~Yan, ``Keyword search using
  attention-based end-to-end asr and frame-synchronous phoneme alignments,''
  \emph{IEEE/ACM Transactions on Audio, Speech, and Language Processing},
  vol.~29, pp. 3202--3215, 2021.

\bibitem{guo2021far}
Y.~Guo, Y.~Chen, G.~Cheng, P.~Zhang, and Y.~Yan, ``Far-field speech recognition
  based on complex-valued neural networks and inter-frame similarity difference
  method,'' in \emph{2021 IEEE Automatic Speech Recognition and Understanding
  Workshop (ASRU)}.\hskip 1em plus 0.5em minus 0.4em\relax IEEE, 2021, pp.
  1003--1010.

\bibitem{Dong2018SpeechTransformerAN}
L.~Dong, S.~Xu, and B.~Xu, ``Speech-transformer: A no-recurrence
  sequence-to-sequence model for speech recognition,'' \emph{Proc. ICASSP}, pp.
  5884--5888, 2018.

\bibitem{mostefa2007chil}
D.~Mostefa, N.~Moreau, K.~Choukri, G.~Potamianos, S.~M. Chu, A.~Tyagi, J.~R.
  Casas, J.~Turmo, L.~Cristoforetti, F.~Tobia \emph{et~al.}, ``The chil
  audiovisual corpus for lecture and meeting analysis inside smart rooms,''
  \emph{Language resources and evaluation}, vol.~41, no.~3, pp. 389--407, 2007.

\bibitem{ami}
S.~Renals, T.~Hain, and H.~Bourlard, ``Recognition and understanding of
  meetings the ami and amida projects,'' in \emph{2007 IEEE Workshop on
  Automatic Speech Recognition Understanding (ASRU)}, 2007, pp. 238--247.

\bibitem{fu2021aishell}
Y.~Fu, L.~Cheng, S.~Lv, Y.~Jv, Y.~Kong, Z.~Chen, Y.~Hu, L.~Xie, J.~Wu, H.~Bu
  \emph{et~al.}, ``Aishell-4: An open source dataset for speech enhancement,
  separation, recognition and speaker diarization in conference scenario,''
  \emph{arXiv preprint arXiv:2104.03603}, 2021.

\bibitem{watanabe2020chime}
S.~Watanabe, M.~Mandel, J.~Barker, E.~Vincent, A.~Arora, X.~Chang,
  S.~Khudanpur, V.~Manohar, D.~Povey, D.~Raj \emph{et~al.}, ``Chime-6
  challenge: Tackling multispeaker speech recognition for unsegmented
  recordings,'' \emph{arXiv preprint arXiv:2004.09249}, 2020.

\bibitem{godfrey1992switchboard}
J.~J. Godfrey, E.~C. Holliman, and J.~McDaniel, ``Switchboard: Telephone speech
  corpus for research and development,'' in \emph{Acoustics, Speech, and Signal
  Processing, IEEE International Conference on}, vol.~1.\hskip 1em plus 0.5em
  minus 0.4em\relax IEEE Computer Society, 1992, pp. 517--520.

\bibitem{liu2006hkust}
Y.~Liu, P.~Fung, Y.~Yang, C.~Cieri, S.~Huang, and D.~Graff, ``Hkust/mts: A very
  large scale mandarin telephone speech corpus,'' in \emph{International
  Symposium on Chinese Spoken Language Processing}.\hskip 1em plus 0.5em minus
  0.4em\relax Springer, 2006, pp. 724--735.

\bibitem{ryant2018first}
N.~Ryant, K.~Church, C.~Cieri, A.~Cristia, J.~Du, S.~Ganapathy, and
  M.~Liberman, ``First dihard challenge evaluation plan,'' \emph{2018, tech.
  Rep.}, 2018.

\bibitem{ryant2019second}
N.~Ryant, K.~Church, C.~Cieri, A.~Cristia, J.~Du, S.~Ganapathy, and
  M.~Liberman, ``The second dihard diarization challenge: Dataset, task, and
  baselines,'' \emph{arXiv preprint arXiv:1906.07839}, 2019.

\bibitem{snyder2018x}
D.~Snyder, D.~Garcia-Romero, G.~Sell, D.~Povey, and S.~Khudanpur, ``X-vectors:
  Robust dnn embeddings for speaker recognition,'' in \emph{2018 IEEE
  international conference on acoustics, speech and signal processing
  (ICASSP)}.\hskip 1em plus 0.5em minus 0.4em\relax IEEE, 2018, pp. 5329--5333.

\bibitem{fox2007sticky}
E.~B. Fox, E.~B. Sudderth, M.~I. Jordan, and A.~S. Willsky, ``The sticky
  hdp-hmm: Bayesian nonparametric hidden markov models with persistent
  states,'' \emph{Arxiv preprint}, 2007.

\bibitem{diez2019analysis}
M.~Diez, L.~Burget, F.~Landini, and J.~{\v{C}}ernock{\`y}, ``Analysis of
  speaker diarization based on bayesian hmm with eigenvoice priors,''
  \emph{IEEE/ACM Transactions on Audio, Speech, and Language Processing},
  vol.~28, pp. 355--368, 2019.

\bibitem{wang2020speaker}
J.~Wang, X.~Xiao, J.~Wu, R.~Ramamurthy, F.~Rudzicz, and M.~Brudno, ``Speaker
  attribution with voice profiles by graph-based semi-supervised learning,''
  \emph{Proc. Interspeech 2020}, pp. 289--293, 2020.

\bibitem{park2020speaker}
T.~J. Park, K.~J. Han, J.~Huang, X.~He, B.~Zhou, P.~Georgiou, and S.~Narayanan,
  ``Speaker diarization with lexical information,'' \emph{arXiv preprint
  arXiv:2004.06756}, 2020.

\bibitem{shafey2019joint}
L.~E. Shafey, H.~Soltau, and I.~Shafran, ``Joint speech recognition and speaker
  diarization via sequence transduction,'' \emph{arXiv preprint
  arXiv:1907.05337}, 2019.

\bibitem{chen2022interrelate}
Y.~Chen, Y.~Guo, Q.~Li, G.~Cheng, P.~Zhang, and Y.~Yan, ``Interrelate training
  and searching: A unified online clustering framework for speaker
  diarization,'' \emph{arXiv preprint arXiv:2206.13760}, 2022.

\bibitem{fujita2019end}
Y.~Fujita, N.~Kanda, S.~Horiguchi, Y.~Xue, K.~Nagamatsu, and S.~Watanabe,
  ``End-to-end neural speaker diarization with self-attention,'' in \emph{2019
  IEEE Automatic Speech Recognition and Understanding Workshop (ASRU)}.\hskip
  1em plus 0.5em minus 0.4em\relax IEEE, 2019, pp. 296--303.

\bibitem{medennikov2020target}
I.~Medennikov, M.~Korenevsky, T.~Prisyach, Y.~Khokhlov, M.~Korenevskaya,
  I.~Sorokin, T.~Timofeeva, A.~Mitrofanov, A.~Andrusenko, I.~Podluzhny
  \emph{et~al.}, ``Target-speaker voice activity detection: A novel approach
  for multi-speaker diarization in a dinner party scenario,'' \emph{Proc.
  Interspeech 2020}, pp. 274--278, 2020.

\bibitem{zhang2019fully}
A.~Zhang, Q.~Wang, Z.~Zhu, J.~Paisley, and C.~Wang, ``Fully supervised speaker
  diarization,'' in \emph{ICASSP 2019-2019 IEEE International Conference on
  Acoustics, Speech and Signal Processing (ICASSP)}.\hskip 1em plus 0.5em minus
  0.4em\relax IEEE, 2019, pp. 6301--6305.

\bibitem{maiti2021end}
S.~Maiti, H.~Erdogan, K.~Wilson, S.~Wisdom, S.~Watanabe, and J.~R. Hershey,
  ``End-to-end diarization for variable number of speakers with local-global
  networks and discriminative speaker embeddings,'' in \emph{ICASSP 2021-2021
  IEEE International Conference on Acoustics, Speech and Signal Processing
  (ICASSP)}.\hskip 1em plus 0.5em minus 0.4em\relax IEEE, 2021, pp. 7183--7187.

\bibitem{horiguchi2020end}
S.~Horiguchi, Y.~Fujita, S.~Watanabe, Y.~Xue, and K.~Nagamatsu, ``End-to-end
  speaker diarization for an unknown number of speakers with encoder-decoder
  based attractors,'' \emph{Proc. Interspeech 2020}, pp. 269--273, 2020.

\bibitem{xue2021online}
Y.~Xue, S.~Horiguchi, Y.~Fujita, S.~Watanabe, P.~Garc{\'\i}a, and K.~Nagamatsu,
  ``Online end-to-end neural diarization with speaker-tracing buffer,'' in
  \emph{2021 IEEE Spoken Language Technology Workshop (SLT)}.\hskip 1em plus
  0.5em minus 0.4em\relax IEEE, 2021, pp. 841--848.

\bibitem{huang2020speaker}
Z.~Huang, S.~Watanabe, Y.~Fujita, P.~Garc{\'\i}a, Y.~Shao, D.~Povey, and
  S.~Khudanpur, ``Speaker diarization with region proposal network,'' in
  \emph{ICASSP 2020-2020 IEEE International Conference on Acoustics, Speech and
  Signal Processing (ICASSP)}.\hskip 1em plus 0.5em minus 0.4em\relax IEEE,
  2020, pp. 6514--6518.

\bibitem{li2022cn}
L.~Li, R.~Liu, J.~Kang, Y.~Fan, H.~Cui, Y.~Cai, R.~Vipperla, T.~F. Zheng, and
  D.~Wang, ``Cn-celeb: multi-genre speaker recognition,'' \emph{Speech
  Communication}, 2022.

\bibitem{chung2018voxceleb2}
J.~S. Chung, A.~Nagrani, and A.~Zisserman, ``Voxceleb2: Deep speaker
  recognition,'' \emph{Proc. Interspeech 2018}, pp. 1086--1090, 2018.

\bibitem{landini2022bayesian}
F.~Landini, J.~Profant, M.~Diez, and L.~Burget, ``Bayesian hmm clustering of
  x-vector sequences (vbx) in speaker diarization: theory, implementation and
  analysis on standard tasks,'' \emph{Computer Speech \& Language}, vol.~71, p.
  101254, 2022.

\end{thebibliography}

\end{document}